%% file: main.tex
\documentclass{article}

\usepackage{RobustML_iclr2021_conference}
\iclrfinalcopy
\usepackage{times}

\input{math_commands.tex}

\usepackage[utf8]{inputenc} 
\usepackage[T1]{fontenc}    
\usepackage{hyperref}       
\usepackage{url}            
\usepackage{booktabs}       
\usepackage{amsfonts}       
\usepackage{nicefrac}       
\usepackage{microtype}      
\usepackage[ruled,vlined, linesnumbered]{algorithm2e} 
\usepackage{todonotes, comment}
\usepackage{amsmath, amsthm}
\usepackage{subcaption}

\newtheorem{thm}{Theorem}[section]
\newtheorem{lem}[thm]{Lemma}

\newtheorem{assum}{Assumption}

\title{Learning Robust Controllers via \\ Probabilistic Model-Based Policy Search}
\author{Valentin Charvet, Bj{\o}rn Sand Jensen, and Roderick Murray-Smith   \\ 
    Department of Computing Science \\ 
    University of Glasgow \\
    \texttt{v.charvet.1@research.gla.ac.uk}
}

\begin{document}

\maketitle

\begin{abstract}
Model-based Reinforcement Learning estimates the true environment through a world model in order to approximate the optimal policy. This family of algorithms usually benefits from better sample efficiency than their model-free counterparts. We investigate whether controllers learned in such a way are robust and able to generalize under small perturbations of the environment. Our work is inspired by the PILCO algorithm, a method for probabilistic policy search. We show that enforcing a lower bound to the likelihood noise in the Gaussian Process dynamics model regularizes the policy updates and yields more robust controllers. We demonstrate the empirical benefits of our method in a simulation benchmark.

\end{abstract}

\section{Introduction}
\label{sec:intro}
\input{Introduction/intro_v0}

\section{Related Work}
\label{sec:related_work}
\input{Related_Work/related_work_v0}

\section{Methods}
\label{sec:methods}
\input{Methods/methods_v1}

\section{Experiments}
\label{sec:experiments}
\input{Experiments/experiments_v0}

\section{Conclusion}
In this paper, we proposed a method to increase the robustness of controllers trained with Model-Based Reinforcement Learning by using a regularization on the Gaussian Process noise hyperparameter. We empirically demonstrated a significant decrease in performance fluctuations for higher noise bounds, hence improving domain adaptation. We also believe these ideas can be applied to the offline RL case and plan to investigate this direction in future work.

\vfill
\pagebreak

\subsection*{Acknowledgements}
This work is supported by EPSRC Grant EP/R018634/1 (Closed-Loop Data Science for Complex, Computationally- and Data-Intensive Analytics) and the University of Glasgow.
We thank the contributors open-source software libraries we used for this work, specifically GPyTorch \citep{gardner2018gpytorch}, PyBullet Gym \citep{benelot2018} and Seaborn \citep{waskom2020seaborn}.

\bibliography{bibliography}
\bibliographystyle{iclr2021_conference}

\vfill
\pagebreak

\appendix
\section{Proof: bound on physical parameters}
\label{app:bound_physics}
\input{Apppendix/bound_disturbance}

\section{Details of the experimental setup}
\label{app:experiments}
\input{Apppendix/experiments}
\end{document}

%% file: math_commands.tex
\usepackage{amsmath,amsfonts,bm}

\def\eqref#1{equation~\ref{#1}}

\def\1{\bm{1}}

\def\rvz{{\mathbf{z}}}

\def\vtheta{{\bm{\theta}}}
\def\vphi{{\bm{\phi}}}   
\def\vdelta{{\bm{\delta}}} 

\def\vu{{\bm{u}}}

\def\vx{{\bm{x}}}

\def\vz{{\bm{z}}}

\DeclareMathAlphabet{\mathsfit}{\encodingdefault}{\sfdefault}{m}{sl}
\SetMathAlphabet{\mathsfit}{bold}{\encodingdefault}{\sfdefault}{bx}{n}

\newcommand{\E}{\mathbb{E}}

\DeclareMathOperator*{\argmin}{arg\,min}

%% file: Introduction/intro_v0.tex

Despite impressive ability at solving tasks such as video games \citep{mnih2013playing}, \emph{Reinforcement Learning (RL)} often fails to work well in real world situations.   
\citet{Dulac-Arnold2019} survey the causes for it and propose a set of nine challenges representing the difficulty to deploy agents outside of simulations. Among them is the problem of non-stationarity, which means the characteristics of the environment evolve during time. It is a similar issue to that of distribution shift, because the agent has not been trained in the new environment it interacts with. Interestingly, this problem can also be found in Batch RL where the policy is learned from a dataset of observations. We refer to the survey \citep{Levine2020} for more details on that.

Learning robust policies is therefore a crucial step for deploying reinforcement learning agents. The reason is that over time, a robot's sensors can suffer from physical degradation which will impair their ability to relay the true state of the environment. Instead, they will only receive noisy estimations of it. Moreover, wear and tear are likely to modify the intrinsic physical nature of the agent or the environment itself. As a consequence, it is necessary to account for this variability during training. 


The common ways to approach that problem include meta and adversarial learning. While the former learns a prior on predictive model from multiple environments the latter optimizes the policy for a pessimistic objective function. 

The method we propose instead only requires one environment to be trained on and does not need to modify the objective function. We choose to build on PILCO by \citet{Deisenroth2011} because its use of probabilistic policy search shows natural robustness. We further regularize the Gaussian Process dynamics in order to increase its domain adaptation capabilities.A similar approach was applied by \citet{igl2019generalization} to deep neural networks. Finally we also integrate safety constraints to evaluate the robustness of our controllers more precisely.



%% file: Related_Work/related_work_v0.tex

The problem of domain adaptation is not exclusive to sequential decision-making problems and arises when training and evaluation datasets are not independents and identically distributed. One of the most classical approach to tackle this is by regularization (\citet{Hastie2009} chapters 3 to 6), a method that improves generalization by preventing overfitting. These tricks work well on both vision \citep{jeong2020consistency, 10.5555/3326943.3327036} and language \citep{10.5555/3326943.3327036} tasks. 

A promising direction of research is that of \emph{meta-learning}. In \citep{Saemundsson2018, nagabandi2018learning}, a prior on the system dynamics is learned during the initial training phase. When running in a new environment, this prior is updated so as to best reflect the new system's dynamics. This type of problem can also be solved by learning a hierarchical policy \citep{kupcsik2013data}, where the upper layer depends on the context and the lower one on the current state of the robot. Similarly \citep{yu2017preparing} learns a unique policy in a several environment and use an identification method that feeds a context signal to the policy. The drawback of these methods is that they require access to a wide range of environment in order to learn a latent representation of the dynamics. In contrast, our method only needs access to one simulator.

Last, robustness can be achieved by optimizing a \emph{pessimistic objective}. This is often referred to as the \textit{Robust Markov Decision Process} (MDP) framework, which can be solved by approximate dynamics programming \citet{mankowitz2018learning, pmlr-v32-tamar14} or within Maximum a Posteriori Policy Optimization \cite{mankowitz2019robust}. Such methods go as back as 2005 \citep{morimoto2005robust} where the authors apply an actor-critic where the controller attempts to correct for disturbances generated by an internal agent. More recently \citep{pinto2017robust} apply a similar method with neural networks. In essence,  these methods solve a minimax optimization problem to account for worst-case scenarios.


%% file: Methods/methods_v1.tex
\subsection*{Model-Based Policy Search}
We consider the setting of an MDP $(X, U, f, C, T)$ denoting respectively the state and action spaces, transition function, cost function and episode length T. The goal is, given a parametric \emph{policy} (or \emph{controller}) $\pi_\vtheta$, to minimize the \emph{return} or \emph{cost-to-go} defined as the expected sum of future costs.
In contrast, model-based RL optimizes the policy on the surrogate objective (\ref{eq:objective_surr}). In that equation, $\hat f$ is the estimate of $f$. It trained by minimizing the empirical risk between the samples $\hat f (\vx_t^i, \vu_t^i)$ and $\vx_{t+1}^i$ encountered in the course of agent-environment interactions.
\begin{equation} 
    \vtheta^* =  
    \underset{\vtheta}{\text{argmin}} ~ \E_{\pi_\vtheta, \hat f} \left[ \sum\nolimits_{t=0}^{T-1} C(\vx_t) \right]. 
\label{eq:objective_surr}
\end{equation}
PILCO \citep{Deisenroth2011} is a model-based algorithm that updates its policy by predicting an approximate trajectory distribution over the episode length $T$. As such, it can be viewed as a model-based Monte-Carlo Policy Gradient method.


The core of the algorithm is the computation of the successor state distribution as $p(\vx_{t+1}) = \int \hat f (\vx_{t+1}| \vx_t, \vu_t)p(\vx_t, \vu_t) d\vx_t d\vu_t$ which is generally an intractable integral. It can be approximated via Moment Matching \cite{girard2003gaussian}, sampling  \citep{Gal2016, Parmas2018} or Numerical Quadrature \citep{Vinogradska2020}. The same method can then be used to approximate the local costs $c_t = \int C(\vx_t) p(\vx_t)d\vx_t$, provided the cost function is in the polynomial or exponential family. 




\subsection*{Learning with Safety Constraints}
In many situations, safety is of critical importance and needs to be incorporated in the policy search. We follow the same procedure as \citet{polymenakos2019safe} but frame it as a \emph{Primal-Dual} problem. This formulation has also been proposed in \citet{cowenrivers2020samba} but its impact on robustness has not been evaluated. Let us consider a set of hazardous regions $H$ that the agent must avoid. We consider the probability of being in that region a \emph{chance constraint} written $G(\rvz) = \int_H p_\rvz(\vz)d\vz$. 

The state-distribution approximation from the previous section (Moment Matching in our case) can be used to integrate the risk over the entire trajectory by computing $G(\vx_t)$ for $t=0 \dots T-1$. For a given threshold $\xi$ we can then write the constrained objective as
\begin{equation} 
\begin{aligned}
    \vtheta^* =  ~
    & \underset{\vtheta}{\argmin}~& \E_{\pi_\vtheta, \hat f} \left[ \sum\nolimits_{t=0}^{T-1} C(\vx_t) \right]
    \,\,\, \text{subject to} \,\,\, ~&  \E_{\pi_\vtheta, \hat f} \left[\sum\nolimits_{t=0}^{T-1}G(\vx_t) \right] < \xi,
\end{aligned}
\label{eq:objective_surr_constraint}
\end{equation}

and the corresponding Lagrangian as
\begin{equation}
    L(\vtheta, \lambda) = V_\vtheta(\vx^0) + \lambda (Q_\vtheta(\vx^0) - \xi),
\label{eq:lagrangian}
\end{equation}
where $V_\vtheta$ and $Q_\vtheta$ are respectively the expected cost and risk-to-go from starting point $\vx^0$.  We can therefore find an optimal solution to the problem (\ref{eq:objective_surr_constraint}) by iteratively  ascending the Lagrangian in $\lambda$ and descending in $\vtheta$. 

\subsection*{Robustness to Perturbations}
The strength of Gaussian Processes for planning is their ability to accurately model predictive uncertainty. However, GPs are not inherently robust to shift in the transition model thus predictions will not be valid when the evaluation environment differs from the training environment. We can mitigate this by forcing the model to be overly cautious by applying a lower bound on the inferred likelihood noise. Intuitively, if we force the learned noise to be greater than some value $\displaystyle \sigma_{low}$ then the policy should be robust in an environment that has observational noise up to a level $\displaystyle \sigma_{low}$. The following simple analysis shows how it helps to generalize for small perturbations of latent parameters of the environment dynamics.



\begin{assum}[Smooth dynamics]
\label{assum1}
We assume the true transition kernel of the system is a smooth function of its latent physical parameters $\vphi$ such that $\vx' = f(\vx, \vu; \vphi)$ and
\begin{equation}
    \|f(\vx, \vu; \vphi) - f(\vx, \vu; \vphi + \bm{\delta})  \| \leq K \| \bm{\delta} \|, \hspace{1cm} \forall \vx, \vu, \vphi, \bm{\delta}
\label{eq:lipschitz_dynamics}
\end{equation}
where $K$ is the Lipschitz constant and $\bm{\delta}$ is a perturbation. 
\end{assum}

\begin{figure}[b!]
    \centering
    \includegraphics[width=\textwidth]{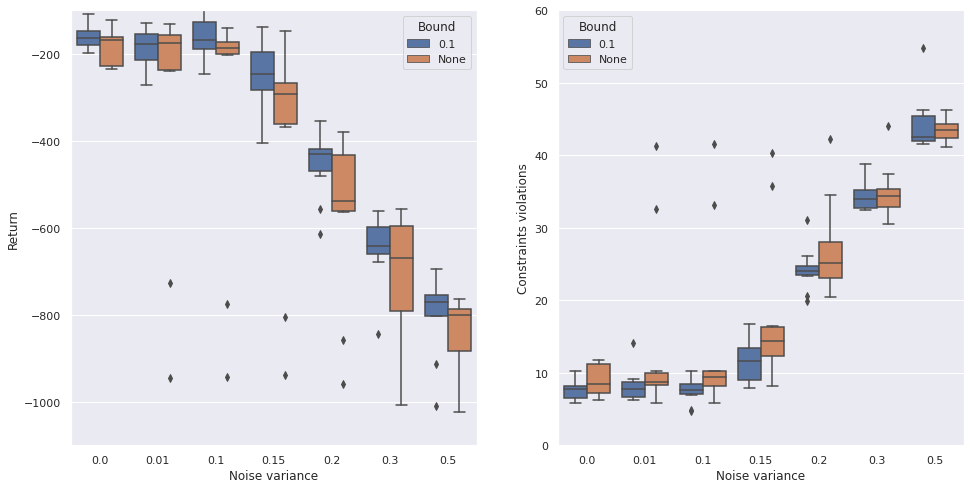}
    \caption{Return (left) and risk (right) for $\sigma_{low}=0.1$ and with no bound. The dots represent values outside the inter-quartiles range (see figure in appendix \ref{app:experiments} for the full graph).}
    \label{fig:performance}
\end{figure}

\begin{assum}
\label{assum2}
The evaluation environment is a perturbed version of the trained one, it transitions according to
$\vx' = f(\vx, \vu; \vphi_0 + \bm\delta )$,
for a given $\bm\delta$ and where $\vphi_0$ are the parameters of the training environment.
\end{assum}

\begin{lem}[Bound on physical perturbations]
\label{eq:physical_variation_bound}
Under assumptions \ref{assum1} and \ref{assum2}, enforcing a lower bound on the noise of the GP model enables optimizing the controller for all the environments in a ball centered at $\vphi_0$ of radius $\bm{\delta}_{\text{max}}$, with $\displaystyle \|\bm{\delta}_{\text{max}}\| \leq \sigma_{low} / {K}$.
\end{lem}
\vspace{-1em}
\begin{proof}
The detail of the derivation is available in appendix \ref{app:bound_physics}.
\end{proof}

The first conclusion we draw from lemma (\ref{eq:physical_variation_bound}) is that the algorithm should be able to generalize in environments similar enough to the training one. The lower $K$ is, the easier it is to generalize. This behavior is what we would expect intuitively, as a low value of $K$ means the environment dynamics vary little with its latent parameters. 
In theory, Lemma (\ref{eq:physical_variation_bound}) means we can generalize in a region with any radius we want by choosing a high value for $\sigma_\text{lbound}$. However, because PILCO predictions are autoregressive, a high value of the noise bound will lead to high variance in the predicted trajectory. This can be alleviated in practice by rolling out the state predictions on short horizons, similarly as in \citet{Janner2019}. In practice, the constant $K$ can be estimated with classical tools \citep{wood1996estimation} but it requires access to versions of the system with different latent parameters. Alternatively to the soft constraint we propose, a prior on the hyperparameter as in \citet{Burnaev2016} acting as a regularization of the evidence maximisation procedure could improve robustness of the policy search. We leave investigation of such priors to future work.


%% file: Experiments/experiments_v0.tex
The question the experiments aim to answer is whether a lower bound of the noise is a good enough regularization to learn robust policies. The experimental setup and hyperparameters are detailed in appendix \ref{app:experiments}.
To evaluate the robustness of our controller to environmental perturbations, we first train the controller on the noise-free environment. We then roll out the controller in the perturbed evaluation environment and record the average returns and risks in 10 runs. We perturb the environment with observation noise as well as variation of the system's parameters, with global variance of perturbation $\sigma_{perturb}$. We show the evaluation performance for a given bound in Figure \ref{fig:performance}. All the results are obtained with models trained from 10 different random seeds.


Table \ref{tab:experiment_summary_pend} summarizes the experiment results, with PILCO without safety constraints as baseline. The first row is obtained with basic PILCO algorithm without safety constraint. We achieve more consistent results with higher values of the bound $0.1$ and $0.2$ hence showing the controllers are more robust. Moreover the values with a bound of $0.1$ and $0.2$ have much less variance than the rest. This means the said models have converged for all 10 random initializations. This is confirmed by figure \ref{fig:performance}, where the distributions of the performance with no bound have many more outliers, that is runs that have not successfully converged. We can see on table \ref{tab:experiment_summary_pend} the trade-off between a high bound that improves generalization and a low one that prevent high variance in the predicted trajectories. Therefore, the most significant improvement of our method is variance reduction: a real benefit as it indicates a more stable training. 



\newcommand{\ret}[1]{\textcolor{blue}{#1}}
\newcommand{\saf}[1]{\textcolor{red}{#1}}

\begin{table}[t]
\footnotesize
    \begin{center}
    \begin{tabular}{|l|| c|c|c|c|c|}
        \toprule[0.5pt]
        $\sigma_{low}$ & \multicolumn{5}{c}{$\sigma_{perturb}$ } \\
        & $0$ & $10^{-2}$ & $10^{-1}$ & $0.15$ & $0.2$ \\
        \midrule[0.5pt]
        None & $-302\pm1506$&$-346\pm192$&$-341\pm185$ &$-501\pm263$&$-630\pm157$\\
        (no safety)& $14\pm85$&$14\pm15$&$17\pm12$&$23\pm18$&$31\pm14$\\ \midrule
        None &$-166\pm1736$&{$-174\pm285$}&$-186\pm290$&$-291\pm259$&$-538\pm191$\\
        & $8\pm78$&$8\pm12$&$9\pm12$&$14\pm10$&$25\pm10$\\ \midrule 
        
        
        $10^{-3}$ &$-197\pm1208$&$-212\pm206$&$-209\pm222$&$-379\pm202$&$-572\pm146$\\ 
        &$8\pm64$&$8\pm11$&$9\pm11$&$11\pm9$&$26\pm7$\\ \midrule 
        
        $10^{-2}$&$-213\pm1060$&$-185\pm190$&$-198\pm202$&$-272\pm180$&$-518\pm112$\\
        &\bm{$7\pm5$}&$8\pm9$&$8\pm8$&$12\pm8$&$23\pm4$\\ \midrule 
        
        $0.1$&\bm{$-163\pm26$}&$\bm{-177\pm45}$&\bm{$-167\pm46$}&$-247\pm75$&\bm{$-420\pm76$}\\ 
        &\bm{$7\pm1$}&\bm{$7\pm2$}&\bm{$7\pm1$}&$11\pm3$&$24\pm3$\\ \hline 
        
        $0.2$&$-171\pm50$&$-184\pm34$&$-176\pm144$&\bm{$-235\pm207$}&$-430\pm169$\\ 
        &\bm{$7\pm1$}&\bm{$7\pm1$}&$8\pm6$&\bm{$10\pm9$}&\bm{$22\pm5$}
        \\ \bottomrule
    
    \end{tabular}
    \end{center}
    \vspace{-0.2cm}
\caption{Median and standard deviation values for performance on the pendulum, top row of each cell is the return (higher is better) and bottom row is number of constraints violations (lower is better). We highlight values with good median performance \emph{and} low variance}
\label{tab:experiment_summary_pend}
\end{table}

%% file: Apppendix/bound_disturbance.tex
\begin{proof}
We show the bound in the case where the system is one dimensonal. In the following we omit the dependency of the dynamics on the control signal for clarity.

Suppose we have trained the estimate of the dynamics on the noise-free system $\vphi_0$, and enforced a bound on the learned likelihood noise $\sigma_{low}$. We write the transitions of the true dynamics $f(\vx, \vphi_0)$ and the estimate learned from it as $\hat f(\vx, \vphi_0)$.

Therefore we have 
\begin{equation}
    \hat f(\vx, \vphi_0) \leq f(\vx, \vphi_0) + \sigma_{low} ~, \hspace{1cm } \text{with high probability.}
\label{eq:app_error_approx}
\end{equation}

Applying the Lipschitz continuity assumption stated in Eq. (\ref{eq:lipschitz_dynamics})
\begin{equation}
\begin{aligned}
f(\vx, \vphi_0 + \vdelta)& \leq f(\vx, \vphi_0) + K \|\vdelta\| \\
    & \leq \hat f (\vx, \vphi_0) + K \|\vdelta\| - \sigma_{low} \hspace{1cm} \text{using (\ref{eq:app_error_approx})}
\end{aligned}
\label{eq:app_intermediate}
\end{equation}

We can write a lower bound for $f(\vx, \vphi_0 + \vdelta)$ similarly as Eq. (\ref{eq:app_intermediate}) and therefore
\begin{equation}
    \displaystyle
    \left| f(\vx, \vphi_0 + \vdelta) - \hat f (\vx, \vphi_0) \right| \leq \left| K \|\vdelta\| - \sigma_{low} \right|
\label{eq:app_final}
\end{equation}

The right-hand-side of equation (\ref{eq:app_final}) is a convex function of $\vdelta$ and is minimized and equal to 0 for 
$\vdelta \in \left\{\vdelta',~ K \|\vdelta'\| < \sigma_{low} \right\}$, which concludes the proof.
\end{proof}

%% file: Apppendix/experiments.tex
\subsection*{Hyperparameters and setup}
We use the pendulum benchmark environment for our experiments. We learn the dynamics with a Sparse Gaussian Process model and the policy is parameterized by a nonlinear Radial Basis Function network. We plan long term trajectories with Moment Matching approximations and both policy parameters and dual variable are optimized via Gradient Descent. We chose to extend the classic PILCO algorithm with safety constraints as they are one of the challenges defined in \citep{Dulac-Arnold2019}. 

The hyperparameters of the algorithm are shown  in table \ref{tab:hyperparams}. We use the same learning rates, optimisation algorithm and epochs for updating the controller and model at each episode. The value $\lambda_0$ is the initial value for the dual variable. The constraint we choose essentially prevents the pendulum from swinging-up the mass from its left side. We use a saturating cost function, centered around the target (pendulum at upright position). The parameters of the pendulum we are perturbing are: pole length, mass and gravity vector.

\begin{table}[ht]
    \caption{Hyperparameters Summary}
    \label{tab:hyperparams}
    \centering
    \begin{tabular}{c|c}

    \textbf{Parameter}& \textbf{Value}  \\ \hline \hline
    Inducing Points& $100$  \\ \hline
    Number of basis functions & $50$ \\ \hline
    Prediction Horizon & 40 \\ \hline
    Number of episodes & 20 \\ \hline
    Optimiser & Adam \\ \hline
    Learning Rate & $0.01$ \\ \hline
    Epochs per episode & $100$ \\ \hline 
    Constraint & $\theta \in [45, 135]$ \\  \hline
    $\xi$ & $1$ \\ \hline
    $\lambda_0$ & 20 \\ \hline
    \end{tabular}
\end{table}

\subsection*{Note On Lower Bound Constraints}
In order to impose a lower bound on the inferred noise, we optimize that hyperparameter in a transformed space. 
Therefore we have $\sigma_{raw} \in \mathbb{R}$, optimized so as to minimize the negative log likelihood. For predictions, this value is transformed so as to be greater than $\sigma_{low}$ with
$\sigma_{noise} = SoftPlus(\sigma_{raw}) + \sigma_{low}$

\begin{figure}[h]
 \label{fig:performance_uncut}
    \centering
    \includegraphics[width=\textwidth]{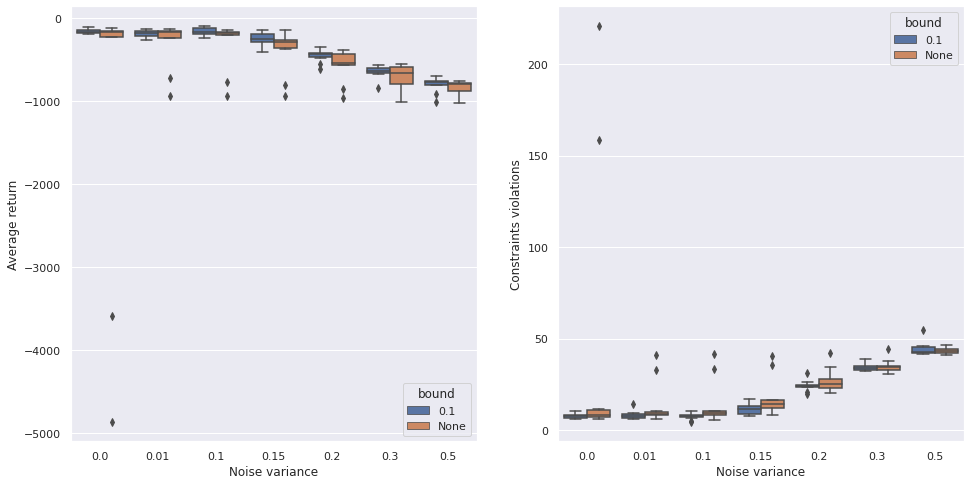}
    \caption{Return (left) and risk (right) with $\sigma_\text{lbound}=0.1$ and no bound. The dots represent values outside the inter-quartiles range}
\end{figure}

